\documentclass[letterpaper, 10 pt, conference]{ieeeconf}  % Comment this line out if you need a4paper

\IEEEoverridecommandlockouts                              % This command is only needed if 
\usepackage{graphics} % for pdf, bitmapped graphics files
\usepackage{epsfig} % for postscript graphics files
\usepackage{mathptmx} % assumes new font selection scheme installed
\usepackage{times} % assumes new font selection scheme installed
\usepackage{amsmath} % assumes amsmath package installed
\usepackage{amssymb}  % assumes amsmath package installed
\usepackage{siunitx}
\usepackage{hyperref}
\UseRawInputEncoding
\usepackage{algorithmic}
\usepackage{graphicx}
\usepackage{textcomp}
\usepackage[capitalise]{cleveref}
\usepackage[dvipsnames]{xcolor}
\usepackage{tabularray} 
\usepackage{subcaption}
\usepackage{acro}
\usepackage{cite}
\usepackage{fancyhdr}
\fancypagestyle{FirstPage}{
	
	\lfoot{{\small \copyright2024 IEEE. Personal use of this material is permitted. Permission from IEEE must be obtained for all other uses, in any current or future media, including
		reprinting/republishing this material for advertising or promotional purposes, creating new
		collective works, for resale or redistribution to servers or lists, or reuse of any copyrighted
		component of this work in other works.}}
}

\DeclareAcronym{BEV}{
	short=BEV, 
	long=Bird's Eye View
}

\DeclareAcronym{GT}{
	short=GT, 
	long=Ground Truth
}

\DeclareAcronym{LR}{
	short=LR, 
	long=Learning Rate
}

\DeclareAcronym{TP}{
	short=TP, 
	long=True Positive
}

\title{\LARGE \bf Improved Single Camera BEV Perception Using Multi-Camera Training
}
\author{Daniel Busch$^{1, 2}$, Ido Freeman$^2$, Richard Meyes$^1$, Tobias Meisen$^1$% <-this % stops a space
\thanks{$^{1}$University of Wuppertal, Germany}%
\thanks{$^{2}$APTIV, daniel.busch@aptiv.com}%
}

\begin{document}

\maketitle
\thispagestyle{empty}
\pagestyle{empty}

%%%%%%%%%%%%%%%%%%%%%%%%%%%%%%%%%%%%%%%%%%%%%%%%%%%%%%%%%%%%%%%%%%%%%%%%%%%%%%%%
\begin{abstract}
\ac{BEV} map prediction is essential for downstream autonomous driving tasks like trajectory prediction. In the past, this was accomplished through the use of a sophisticated sensor configuration that captured a surround view from multiple cameras. However, in large-scale production, cost efficiency is an optimization goal, so that using fewer cameras becomes more relevant. But the consequence of fewer input images correlates with a performance drop. This raises the problem of developing a \ac{BEV} perception model that provides a sufficient performance on a low-cost sensor setup. 
Although, primarily relevant for inference time on production cars, this cost restriction is less problematic on a test vehicle during training. Therefore, the objective of our approach is to reduce the aforementioned performance drop as much as possible using a modern multi-camera surround view model reduced for single-camera inference. 
The approach includes three features, a modern masking technique, a cyclic \ac{LR} schedule, and a feature reconstruction loss for supervising the transition from six-camera inputs to one-camera input during training.
Our method outperforms versions trained strictly with one camera or strictly with six-camera surround view for single-camera inference resulting in reduced hallucination and better quality of the \ac{BEV} map.\\
\end{abstract}


%%%%%%%%%%%%%%%%%%%%%%%%%%%%%%%%%%%%%%%%%%%%%%%%%%%%%%%%%%%%%%%%%%%%%%%%%%%%%%%%
\thispagestyle{FirstPage}

 \begin{figure*}[ht]
	\centering
	\includegraphics[width=0.75\textwidth]{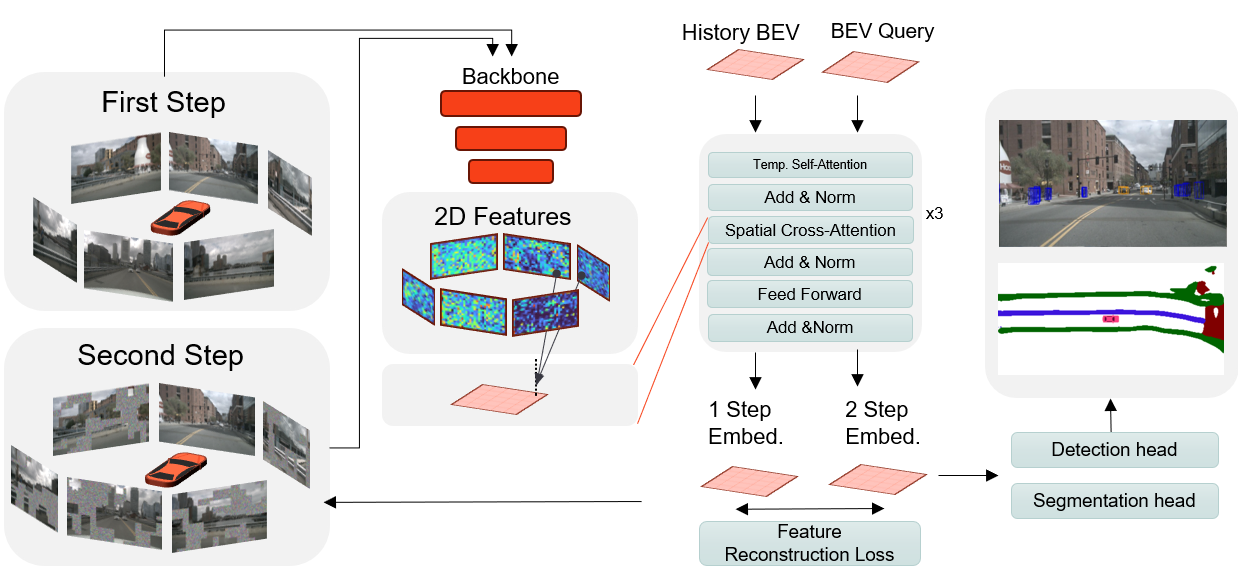}
	\caption{BEVFormer architecture \cite{li_bevformer_2022} extended with the feature reconstruction method. Left: First-step input and second-step input with noise masking. Midsection: Backbone and Transformer layers with Temporal Self-Attention into History \ac{BEV} and Spatial Cross-Attention with re-projection into the 2D features from the backbone. Additionally, the Feature Reconstruction loss over the \ac{BEV} features embeddings from the first and second steps. Right: Heads and output samples.}
	\label{Architecture}
\end{figure*}

\section{INTRODUCTION}

\ac{BEV} map prediction delivers easily interpretable traffic scene information. It implicitly includes objects and their positions in world coordinates. Many modern methods can extract the needed semantic information and predict the \ac{BEV} e.g. \cite{yang_bevformer_2023, pan_cross-view_2020, jiang_polarformer_2023, li_fb-bev_2023}. With the use of such state-of-the-art methods, it is now feasible to generate full scenes from just a few seconds of recorded footage captured by a sophisticated camera setup. However, a problem with these methods for such environmental perception is their need for multiple cameras to cover a $360$ degrees surround view during training and inference. Some even require additional sensors like radar or lidar \cite{harley_simple-bev_2023,li_bev-lgkd_2024}. On the other hand, methods using only a single front camera come with a significant drop in quality. For example in \cite{wang_pseudo-lidar_2019}, a Pseudo-LiDAR model is developed that loses performance along with two benchmark models due to the reduction from stereo to single camera. Moreover, in \cite{li_delving_2024} several different approaches were compared on the nuSecnes dataset \cite{caesar_nuscenes_2020}, with a single camera method performing second worst. This is understandable up to a certain extent, as they receive less input information. Apart from highly equipped research vehicles, the bulk of production vehicles just have a front camera. Even though, some low-volume premium vehicles already have more cameras, adding a comparably low-priced camera will have a large financial impact on higher production volumes. Accordingly, bringing single-camera models as close as possible to the performance of a modern surround-view model is beneficial for mass-production vehicles. As stated in \cite{roddick_predicting_2020} for sufficient perception of the whole scene a multi-camera setup is needed. This also underlines the performance drop by the reduction just from stereo to single camera input reported in \cite{li_delving_2024}.\\
This paper presents a method to reduce the performance drop between training with a full environment view using a multi-camera setup and inference that can be performed with only one camera. The method intelligently reduces the information of the multi-camera setup during the training phase. More precisely, it combines the advantages of BEVFormer \cite{li_bevformer_2022} as a modern surround view model, with a single front camera limitation during inference. In this way, our trained model benefits from the different camera angles of the surround view and handles aspects such as object shadows and occlusion more robustly. To do that, we present the following three contributions: First, we utilized a state-of-the-art masking technique known as inverse block masking \cite{baevski_efficient_2022} from a modern self-monitoring approach. The ratio of this masking is stepwise increased over the training epochs. The increase ends at the limit of the single front view. Additionally, we ignore \ac{GT} bounding boxes in the loss computation if their corresponding input images are completely masked. Secondly, a cyclic \acl{LR} schedule is introduced to align with the masking method. Due to the different masking ratios, the input data distribution changes. Therefore, the \acf{LR} is aligned to enable the model to transition between the changing data distributions. Lastly, the full sample containing all six camera inputs is used to supervise the masked sample. To achieve this, we introduce a BEV feature reconstruction loss that is targeted at the performance of the surround view BEVFormer model. Combining these features, we propose our final training method that increases the performance of the BEVFormer for single-camera inference. Compared to a single camera training, the mIoU of our model has increased by $19\%$ and the mAP by $414\%$. These numbers reflect a better quality in the \ac{BEV} map and a drastic decrease in the number of false positive detections, since the baseline was trained on objects that lie outside the single camera's view.

\section{Related Work}
\subsection{Inputs for single camera \ac{BEV} models}
Depending on the point of view, reducing input information of a surround-view model or adding input information to a single camera model leads to the same approach. Utilizing additional inputs from other cameras, other time steps or even other sensor types for better performance is not new for \ac{BEV} prediction models \cite{pan_cross-view_2020,rashed_bev-modnet_2021,avidan_lidar_2022}. The method in \cite{pan_cross-view_2020} from the robotics domain performs a camera rotation to get a surround-view input instead of utilizing multiple cameras. Moreover, in \cite{saha_translating_2022} an optional dynamics module can exploit additional temporal information by using the same sensor setup. BEV-MODNet \cite{rashed_bev-modnet_2021} exploits two sequential images to improve the 3D detection of moving objects. Besides the utilization of temporal information, the models presented in \cite{wang_pseudo-lidar_2019} show an increase in performance from mono to stereo camera training for 3D object detection. In \cite{roddick_predicting_2020}, they explain the need for a full surround view to perceive a whole traffic scene and provide a method that fuses the \ac{BEV} feature maps from different camera views. In this way, it extended to a full surround view model. However, even though the previous methods benefit from their extended sensor inputs, the setups stay the same for training and inference. In contrast, in LPCG \cite{avidan_lidar_2022}, more inputs are used during training than on inference by introducing a lidar sensor for label guidance. Thus, it benefits from the lidar data but still just needs the single camera setup for inference.

\subsection{Inputs for multi camera \ac{BEV} models}
Instead of reducing inputs in multi-view \ac{BEV} perception models, extending inputs for better performance is often done following the same principle of additional training input: In \cite{han_exploring_2023} and \cite{park_time_2023} long-term temporal fusion strategies are developed to extract more information from past frames. In BEVStereo \cite{li_bevstereo_2023}, a combination of mono and temporal stereo depth estimation is used as an iterative optimization process. In addition, the authors utilize lidar data during training. Lidar is also used in BEV-LGKD \cite{li_bev-lgkd_2024}, a knowledge distillation framework that is extended by lidar guidance for better performance. Furthermore, in BEVDepth lidar is applied for \ac{GT} data \cite{li_bevdepth_2023}. The PETRv2 \cite{liu_petrv2_2023} model extends the base PETR \cite{avidan_petr_2022} model by a history input. Moreover, the time horizon differs for training and inference. During training time, it is sampled flexibly from between 3 and 27 full lidar rotations in the past whereas on inference a sample of 15 rotations in the past is selected. Thus, the model has a greater variety of time horizons and time steps which makes the model more robust for different vehicle speeds. 
The purely camera-based BEVFormer \cite{li_bevformer_2022} does similarly exploit past frames with its temporal self-attention. In addition, the input is extended by an extra time step during training. In total, it uses three random samples from a two-seconds time horizon, whereas during inference, this is reduced to two consecutive samples. 
The above-mentioned methods like \cite{li_bevformer_2022, liu_petrv2_2023, li_bev-lgkd_2024, li_bevdepth_2023} are still considered full surround view methods, but with additional inputs in the form of time steps or lidar inputs that were not considered during inference.

\section{Method}
Our approach is based on the modern BEVFormer \cite{li_bevformer_2022} for predicting a \ac{BEV} map, which we combine with a ResNet50 \cite{he_deep_2016} backbone. To reduce the BEVFormer from a surround view to a single camera inference we combined three approaches: 
\begin{itemize}[]
	\item Firstly, we implement the inverse block masking \cite{baevski_efficient_2022}.
	\item Secondly, we adapt the cyclic \acf{LR} schedule in response to the change in the input data distribution due to different masking ratios.
	\item Lastly, we introduce a loss called \ac{BEV} feature reconstruction loss to rate how well the \ac{BEV} features are reconstructed out of partially masked image parts. 
\end{itemize}

\subsection{Model Architecture}
The BEVFormer architecture is visualized in \cref{Architecture}. It uses two deformable attention mechanisms based on deformable DETR \cite{zhu_deformable_2021}, named spatial cross-attention and temporal self-attention \cite{li_bevformer_2022}. Grid-shaped \ac{BEV} queries are expanded into the vertical dimension by uniformly distributed reference points. These are projected into the 2D image feature maps that are predicted by the CNN backbone. The spatial cross-attention takes place only in the 2D image feature maps into which the point is reprojected and the features are sampled around their corresponding reference point. The temporal attention exploits the history \ac{BEV} features by first aligning them with the current time step to compensate for object motions. Then the self-attention takes place. In total, it has three transformer layers, which corresponds to a mid-size version provided by \cite{bin-ze_bevformer_segmentation_detection_2023}. This version is chosen to reduce the time and computational effort. Afterwards, two heads are added, one detection head responsible for the 3D bounding box prediction and one segmentation head for the \ac{BEV} segmentation of lane markings. 

\subsection{Approach}
\subsubsection{Masking Methods}\label{masking_methods}
\begin{figure}[bp]
	\centering
	\includegraphics[width=.9\linewidth]{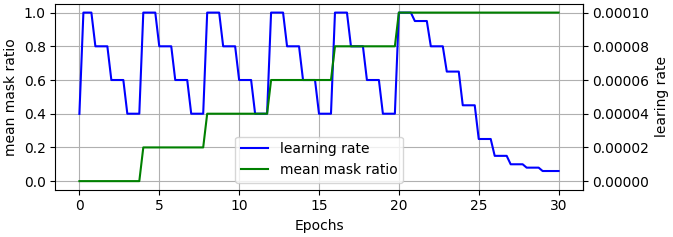}
	\caption{Cyclic \ac{LR} schedule (\textcolor{blue}{blue}) and mean for masking ratio (\textcolor{OliveGreen}{green}) over the training epochs. The masking ratio refers only to the five non-front-facing cameras.}
	\label{LRSchedule}
\end{figure}

\begin{figure}[bp]
	\centering
	\includegraphics[width=.9\linewidth]{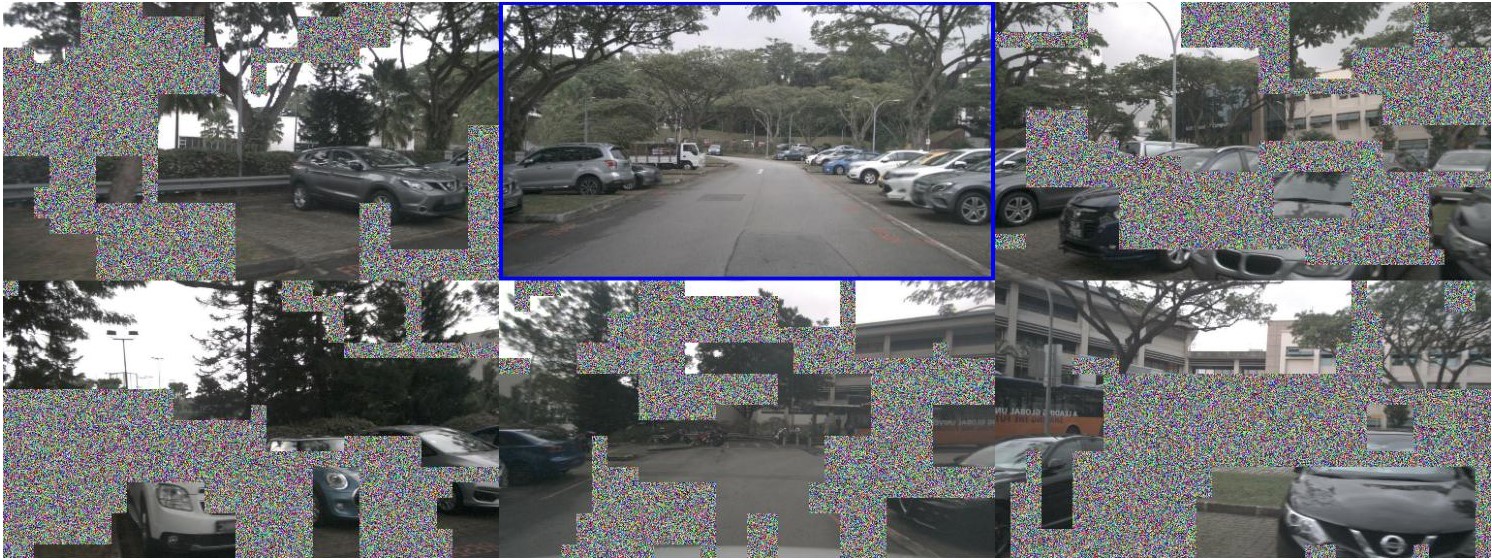}
	\caption{Sample of the inverse block masking with a masking ratio of $\mu=0.4$ and variance $\sigma=0.2$. The front view (\textcolor{blue}{blue} frame) is not masked.}
	\label{MaskingSample}
\end{figure}

The first part of our algorithm relies on the stepwise reduction of usable camera input by using the inverse block masking method \cite{baevski_efficient_2022}. Since we are limiting ourselves to the front camera, the masking is applied only to the five non-front-facing cameras. The step height and width are balanced out such that the input information is reduced only by a small portion ($20\%$) and the network is trained for four epochs before further increasing the masking ratio. Thus, the network can utilize these four epochs to handle the set ratio of missing information by attending to hints from visible portions. Using masks for this purpose is a common practice in self-supervised learning methods as discussed for example in \cite{woo_convnext_2023,li_uniform_2022,oquab_dinov2_2024}. The graph of the mean masking ratio is visualized in \cref{LRSchedule}. To give the masking method more variety during training, the masking ratio is sampled by a Gaussian distribution with a fixed mean ($\mu$) for every reduction step. A masked input sample with a ratio of $\mu=0.4$ is shown in \cref{MaskingSample}. The inverse block masking was originally designed to mask images leaving rectangular contiguous regions visible to provide enough context for a reconstruction of the noised parts. In this way, the model can learn to predict features in hidden regions based on reliable data from visible regions.\\
Additionally, a \ac{GT} bounding box filter is implemented. It filters the \ac{GT} boxes by the camera view angle to force the model to completely neglect blind views produced by the masking method. The \ac{GT} filtering is used during training in the last epochs where the model only receives the front view input. Then, the  \ac{GT} bounding boxes are filtered for all completely blind camera views except for the visible front view. In this context, the front view angle is extended on both sides by a tolerance angle. This tolerance area is just out of view. Thus, history information could still be meaningful as long as the performance metrics will not drop significantly due to further angle extension.

\subsubsection{\ac{LR} Schedule}
The second feature of our approach deals with the adjustment of the \ac{LR}. As described in \cite{smith_cyclical_2017} the \ac{LR} is a crucial hyper-parameter and can slow down the training or even result in divergence of the loss. The BEVFormer uses a cosine annealing \ac{LR} scheme which does not take a change in the data distribution during training into account. Therefore, we align the \ac{LR} with the stepwise increasing masking ratio using the cyclic \ac{LR} scheme depicted in \cref{LRSchedule}. The idea is that at the beginning of every cycle, the \ac{LR} is large enough to give the network the chance to react to the new data distribution. During the cycle, the \ac{LR} is slowly decreased for tuning. During the last epochs at $100\%$ masking ratio, the \ac{LR} is further reduced into small values for fine-tuning.
\subsubsection{Reconstruction Loss}
The third feature of our approach introduces a \ac{BEV} feature reconstruction loss which considers the masked input modified by \ref{masking_methods} as a second sample. The procedure is visualized in \cref{Architecture}.
Each training sample is fed to the network twice. In the first step it is used without any masking and the \ac{BEV} features are kept in memory. The sample is then fed to the network again, now with the mask applied. After the second step, the \ac{BEV} feature reconstruction loss is computed as an L2 loss which is used for a similar purpose in \cite{baevski_efficient_2022}. It is computed between the features obtained with and without masking, constraining the features from masked inputs to be close to the ones from the original input.
\begin{figure*}[t]
	\centering
	\begin{subfigure}{\linewidth}
		\parbox[b][3.5cm][c]{0.05\textwidth}{
			\caption{}
			\label{img:16_baseline1image}
		}
		\parbox[b]{\textwidth}{
			
			\rotatebox[origin=l]{90}{
				%\tiny
			1 Camera Baseline
			}\includegraphics[width=0.88\linewidth]{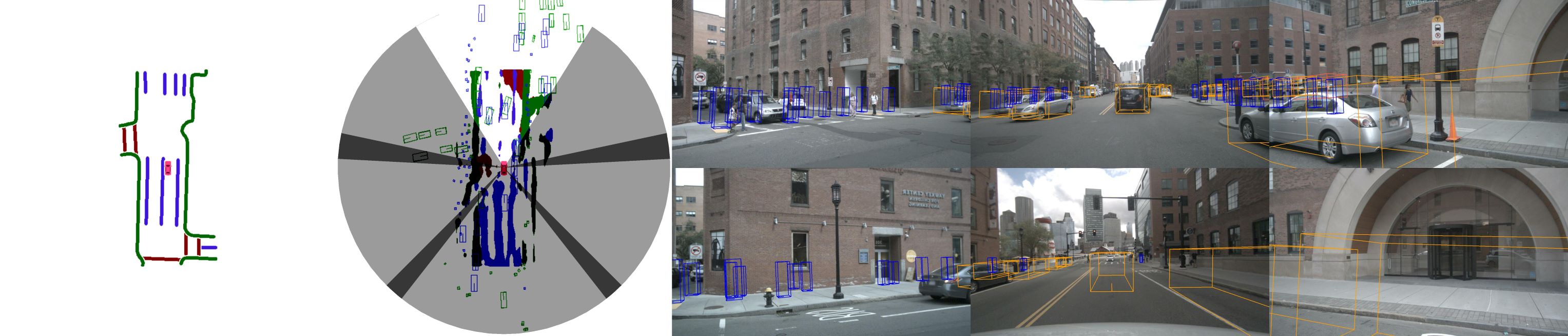}		
		}
	\end{subfigure}
	\begin{subfigure}{\linewidth}
		\parbox[b][3.4cm][c]{0.05\textwidth}{
			\caption{}
			\label{img:16_baseline_all6}
		}
		\parbox[b]{\textwidth}{
			\rotatebox[origin=l]{90}{
			6 Camera Baseline
			}\includegraphics[width=0.88\linewidth]{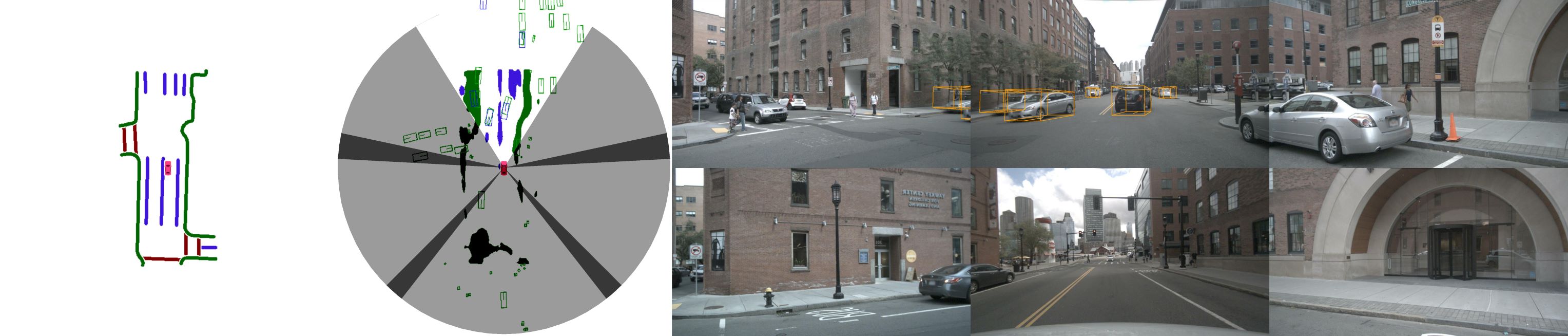}
		}
	\end{subfigure}
	\begin{subfigure}{\linewidth}
		\parbox[b][3.4cm][c]{0.05\textwidth}{
			\caption{}
			\label{img:16_all3}
		}
		\parbox[b]{\textwidth}{
			\rotatebox[origin=l]{90}{
				\ \ \ \ \ \ \ \ \ \ Ours
			}\includegraphics[width=0.88\linewidth]{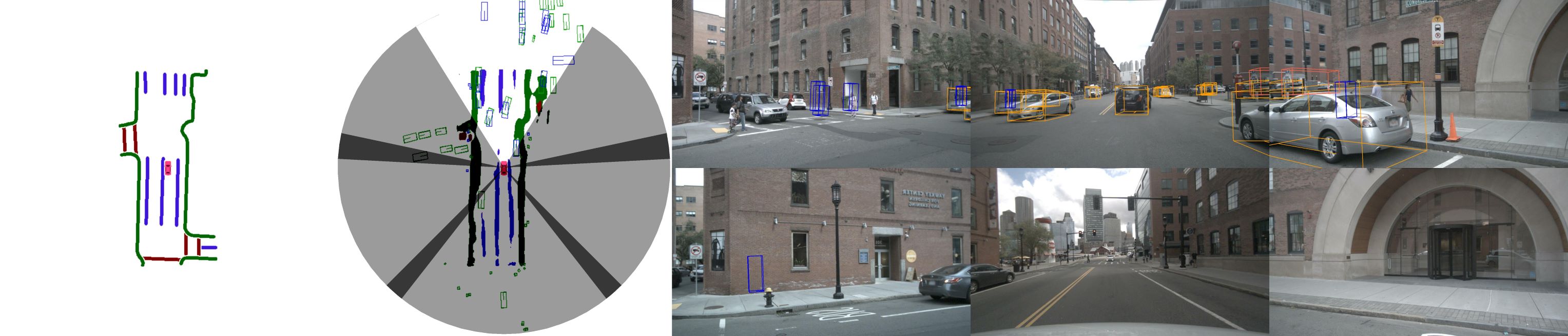}
		}
	\end{subfigure}
	\caption{Results of one sample on two baselines the first one is trained on one camera, the second one is trained on six cameras and results from our method. The inference for all runs is done on one camera. Left: The \ac{GT} segmentation map. Center: The predicted \ac{BEV} map with projected bounding boxes (\ac{GT}=\textcolor{OliveGreen}{green}; prediction=\textcolor{blue}{blue}; masked view=\textcolor{darkgray}{grey}).}
	\label{VisualResults_sample16}
\end{figure*}
\subsection{Dataset}\label{dataset}
The features are trained and tested on the public nuScenes dataset \cite{caesar_nuscenes_2020}. It contains $1000$ traffic scenes of $20s$ in length. The recording vehicles were equipped with one lidar, five radars and a six-camera surround view. It has annotations for 23 object classes as well as HD maps of the road layout around the ego-vehicle \cite{caesar_nuscenes_2020}. The nuScenes developers have defined several validation metrics. To quantify detection quality, they compute the mean average precision (mAP) which is averaged over all classes using \ac{BEV} bounding box center distance for the thresholds. Furthermore, five \ac{TP} scores are defined named as average translation (ATE), scale (ASE), orientation (AOE), velocity (AVE) and attribute (AAE) error. The nuScenes detection score (NDS) takes all previous metrics into account in the following way: $NDS=\frac{1}{10}\left[5mAP+\sum_{mTP\in\mathbb{TP}}(1-min(1,mTP))\right]$ \cite{caesar_nuscenes_2020}. Thus, the mAP is weighed with $50\%$ against the true positive scores. Lastly, the  mean Intersection over Union (mIoU) is used to rate the \ac{BEV} map segmentation. Each metric is computed both as a mean over all classes and individually.

\subsection{Training and Experimental Setup}
A ResNet50 \cite{he_deep_2016} backbone is used and pre-trained on the ImageNet dataset \cite{deng_imagenet_2009}. It is chosen as a trade-off between training time and quality. The model is trained on one A100 GPU for 30 epochs. The implementations are based on BEVFormer as published in \cite{bin-ze_bevformer_segmentation_detection_2023}.\\
Our experiments can be divided into three main sections. Firstly, we evaluate the reduction in false-positive detections for masked image regions achieved by filtering \ac{GT} bounding boxes. To isolate the effect of the \ac{GT} bounding box filter, the model is trained on the front camera only, once with the \ac{GT} filter and once without. In this case, the evaluation is done considering all \ac{GT} boxes of the whole $360$ degrees view to consider also the false positive detections in the camera views that are masked. Secondly, the combination of all three approaches is compared against two baselines: One with a single front camera training and one with the total surround view training. Lastly, a detailed ablation study is done to isolate and compare each approach. For all runs with the inverse block masking technique, the variance of the masking ratio is set to $\sigma=0.2$ except for the first and last cycle where the variance is set to $\sigma=0$. The mean ($\mu$) is stepwise increased in $20\%$ steps as described by \cref{LRSchedule}.
\subsection{Validation}
To focus on the actual effect of our approach, the \ac{GT} bounding boxes are only considered within a $90$ degrees opening angle facing in the driving direction. The camera has an aperture angle of $64.5$ degrees leaving a tolerance angle of $12.75$ degrees to each side. In this area, temporal attention could deliver meaningful output out of history \ac{BEV} features. Therefore, \ac{GT} bounding box filtering is performed everywhere outside the $90$ degrees front facing field-of-view. For comparability, this field of view is consistent for all approaches and baselines.
\section{Results}
\begin{table}[bp]
	\centering
	\resizebox{0.4\textwidth}{!}{
		\begin{tblr}{c|ccc}
			\hline[1pt]
			\ac{GT} bounding box filter & NDS $\uparrow$ & mAP$\uparrow$& mIoU $\uparrow$ \\
			\hline
			\checkmark
			&\textbf{0.1585}
			&\textbf{0.0200}
			&\textbf{0.2063}
			\\
			& 0.1570
			& 0.0187
			& 0.1998
			\\
			%\checkmark & 360° & 121861
			%&\textbf{0.1833}
			%&\textbf{0.0229}
			%&\textbf{0.2173}
			%\\
			%& 360° & 121861
			%& 0.1800
			%& 0.0234
			%& 0.2169
			%\\
			\hline[1pt]
	\end{tblr}}
	\caption{Results of the BEVformer trained on a single camera baseline with and without the \ac{GT} bounding box filter for hallucination suppression. The inference is done using all \ac{GT} bounding boxes of the whole $360$ degrees scene.}
	\label{table:GT_suppress}
\end{table}
\begin{table}[bp]
	\centering
	\resizebox{0.4\textwidth}{!}{
		\begin{tblr}{c|ccc}
			\hline[1pt]
			Method & NDS $\uparrow$ & mAP$\uparrow$& mIoU $\uparrow$ \\
			\hline
			1 camera baseline &0.2081 & 0.0251 & 0.2173   \\
			6 camera baseline & 0.2293 & 0.1024	& 0.1611\\
			all 3 approaches &\textbf{0.2757} &\textbf{0.1290} &\textbf{0.2588} \\
			\hline[1pt]
	\end{tblr}}
	\caption{The combination of the three approaches (Inverse block masking, cyclic \ac{LR} and feature reconstruction loss) compared to six and one camera baseline.}
	\label{all3vsBase}
\end{table}
\begin{table*}[t]
	\centering
	\resizebox{0.88\textwidth}{!}{
		\parbox[c][46mm][c]{0.9\textwidth}{}
		\begin{tblr}{c c c|l l| l l l l l| l}
			\hline[1pt]
			InvBlockMask&CyclicLR& FeatureReconLoss&NDS $\uparrow$&mAP $\uparrow$&mATE $\downarrow$&mASE $\downarrow$&mAOE $\downarrow$&mAVE $\downarrow$&mAAE $\downarrow$&mIoU $\uparrow$ \\[0.5ex]
			\hline
			& &	&0.2293&0.1024&1.0129&0.3387&0.6857&0.9318&0.2622&0.1611\\ 
			\checkmark& &&0.2335&0.0737&0.9646&0.3056&0.6797&0.8152&0.2682&0.2287\\
			& \checkmark & &0.1790 &0.0229 & 1.0884&0.3361&0.6862&1.0035&0.3000&0.2152 \\
			\checkmark&\checkmark & &0.2058 &0.0312 & 1.0753 &0.3269 &0.6554 &0.8693 &0.2464&0.2235\\
			&&\checkmark&0.2460	&0.0761&1.0017&0.2850&0.6305&0.7698 &\textbf{0.2351}&0.2456\\
			\checkmark&&\checkmark&0.2708& 0.0916 &0.9654&\textbf{0.2830}&\textbf{0.5960}&\textbf{0.6651}&0.2407&0.2212\\
			&\checkmark&\checkmark&0.2472&0.1080&0.9611& 0.3047&0.6404&0.9090&0.2534&0.2123\\
			\checkmark&\checkmark&\checkmark& \textbf{0.2757}&\textbf{0.1290}&\textbf{0.9579}&0.2949&0.6161&0.7786&0.2407& \textbf{0.2588}\\
			\hline[1pt]	
	\end{tblr}}
	\caption{Ablation study, for the inverse block masking, cyclic \ac{LR} and feature reconstruction loss. The baseline is trained on 6 cameras. $\uparrow$ higher values are better. $\downarrow$ lower values are better.}
	\label{ablation_study}
\end{table*}
\subsection{\ac{GT} bounding box filter} \label{bbox_filter}
To suppress false-positive detections, we implement the \ac{GT} bounding box filter during training in the last ten epochs where all non-front-facing cameras are fully masked ($100\%$ masking ratio). The effect of this \ac{GT} filter is shown in table \ref{table:GT_suppress}. The mAP score is the most meaningful metric as it is lower for rising false positive values. We observe that all metrics for object detection and semantic segmentation improve.
\subsection{Evaluation of combined features}
The combination of all three features is compared against our baselines in table \ref{all3vsBase}.
One baseline is trained with all six cameras and one baseline is trained only on the front camera.
The evaluation is done only on the front camera with the \ac{GT} bounding box filter applied as described in \cref{masking_methods} for all three runs. Our method outperforms both baselines in the two most important metrics for the object detection NDS and mAP by $20\%$, $25\%$ compared to the second best value. The NDS is a weighted sum of the mAP and the five \ac{TP} scores and the mAP considers false positive values. Additionally, the mIoU is improved by $19\%$ which is the only measured indicator for the semantic segmentation of the \ac{BEV} map.

Apart from the quantitative results, \cref{VisualResults_sample16} shows the results qualitatively on one representative sample. The model trained on one camera (\cref{img:16_baseline1image}) shows the highest false positive rate in the blind areas and the visible front view compared to the other two runs. The semantic segmentation appears also most hallucinative and inaccurate for the single-camera run. Even though it shows many lane and object information in the blind areas it looks less precise and most different to the corresponding \ac{GT} map. For example, the merging street in the left which is just out of view is missing. The baseline trained on six cameras (\cref{img:16_baseline_all6}) looks closer to our approach in the visible front view. Besides that, it only predicts segmentation artifacts in the blind area. Additionally, it provides almost no hint of the semantic segmentation map in the area behind the ego vehicle. Our approach (\cref{img:16_all3}) shows a more accurate \ac{BEV} map also in areas that are just out of view. E.g. it shows the corner of the left intake even though it is not seen anymore by the front view. Moreover, it predicts the highly occluded pedestrian on the left side of view. It shows fewer false-positive detections compared to the single camera baseline but also predicts some information that is out of view.
\subsection{Ablation Study}\label{ablation}
Table. \ref{ablation_study} shows the results of the detailed ablation study. Each feature, including inverse block masking, cyclic \ac{LR}, and feature reconstruction loss, was tested individually as well as in combination to determine their effectiveness. The baseline without any of our features is trained on all six cameras and all runs use the \ac{GT} bounding box filter as tested in \cref{bbox_filter}. Additionally, all runs have only the front view as input information during the inference. Each of the isolated feature runs shows an improvement at least in one metric but on the cost of a decrease in another metric. The isolated feature reconstruction loss shows the most significant improvement in the mIoU. Considering only the NDS, the feature reconstruction loss in combination with the inverse block masking shows the most significant improvement. Besides this, the mAP has the best improvement for the cyclic \ac{LR} in combination with the feature reconstruction loss. However, the combination of all three approaches delivers the best results for the NDS, mAP and mIoU. Additionally, the five true positive errors are among the first three places in their category.

\section{Discussion}
In this paper, we presented our enhanced training method that contains the inverse block masking technique aligned with a cyclic \ac{LR} schedule and a feature reconstruction loss for supervising the transition from six camera training inputs to a single front camera inference. Our method outperforms the two baselines in the important metrics.

\subsection{Effects in latent space}
\begin{figure}[t]
	\centering
	\begin{subfigure}{.23\linewidth}
		\centering
		\includegraphics[width=\linewidth]{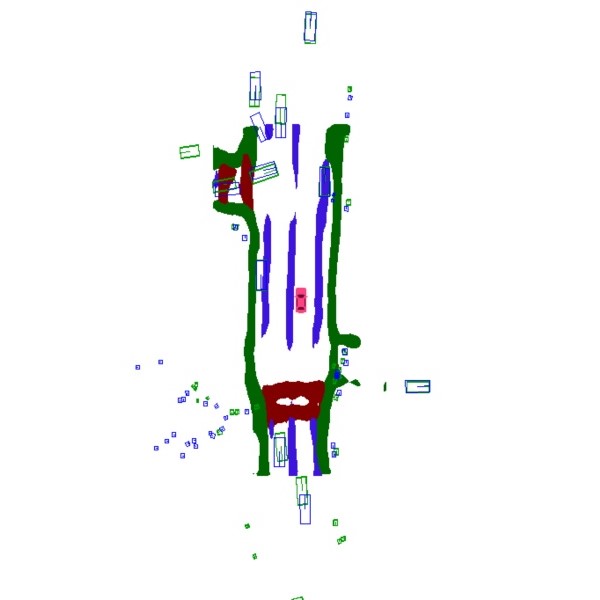}
	\end{subfigure}
	\begin{subfigure}{.23\linewidth}
		\centering
		\includegraphics[width=\linewidth]{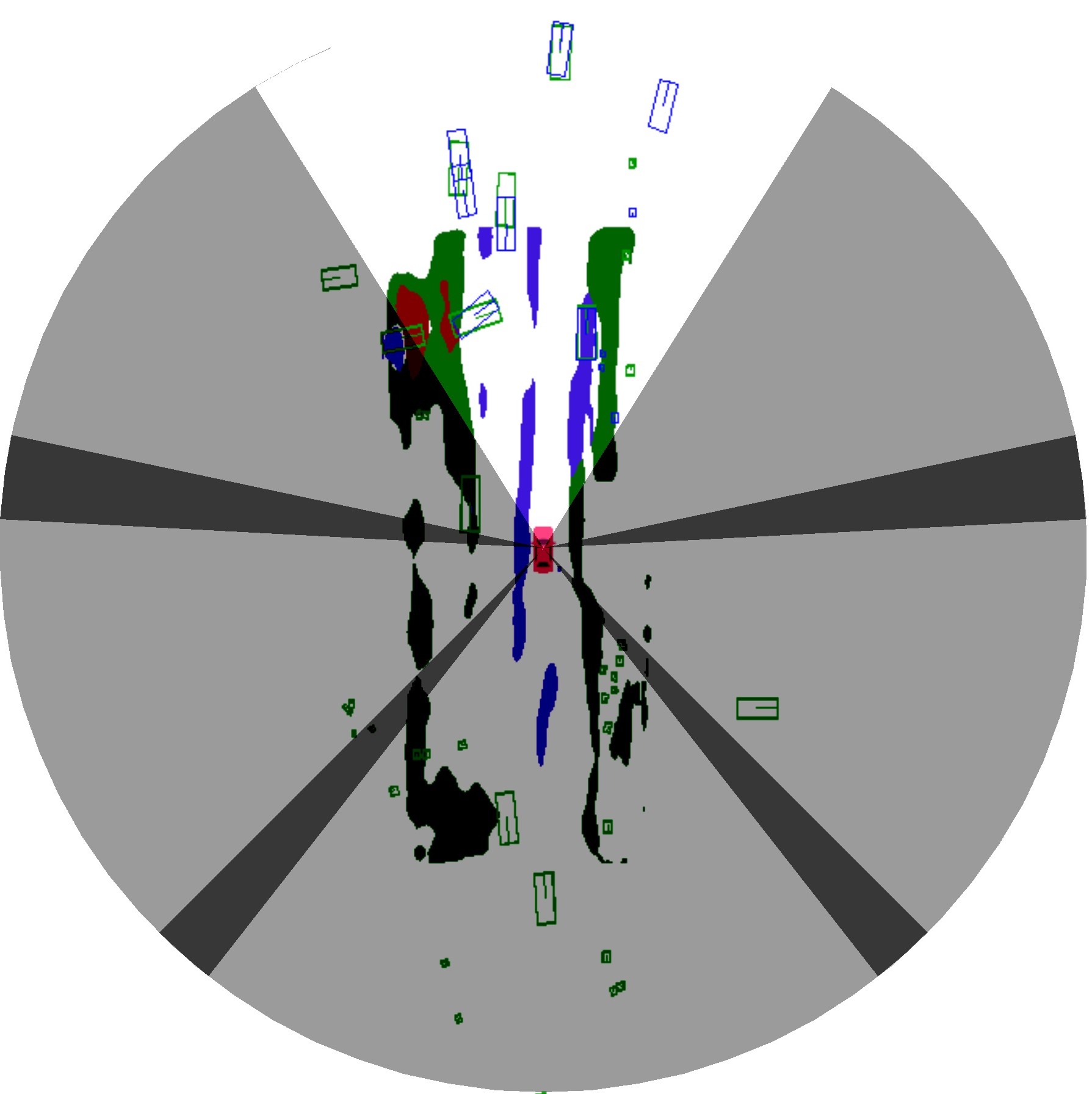}
	\end{subfigure}
	\begin{subfigure}{.23\linewidth}
		\centering
		\includegraphics[width=\linewidth]{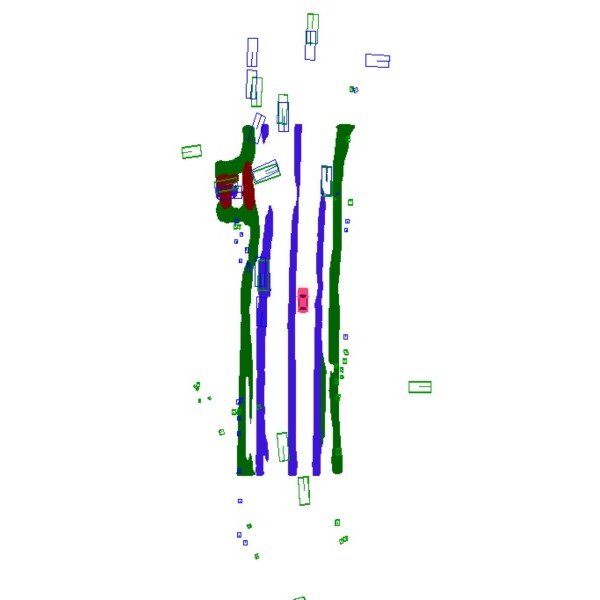}
	\end{subfigure}
	\begin{subfigure}{.23\linewidth}
		\centering
		\includegraphics[width=\linewidth]{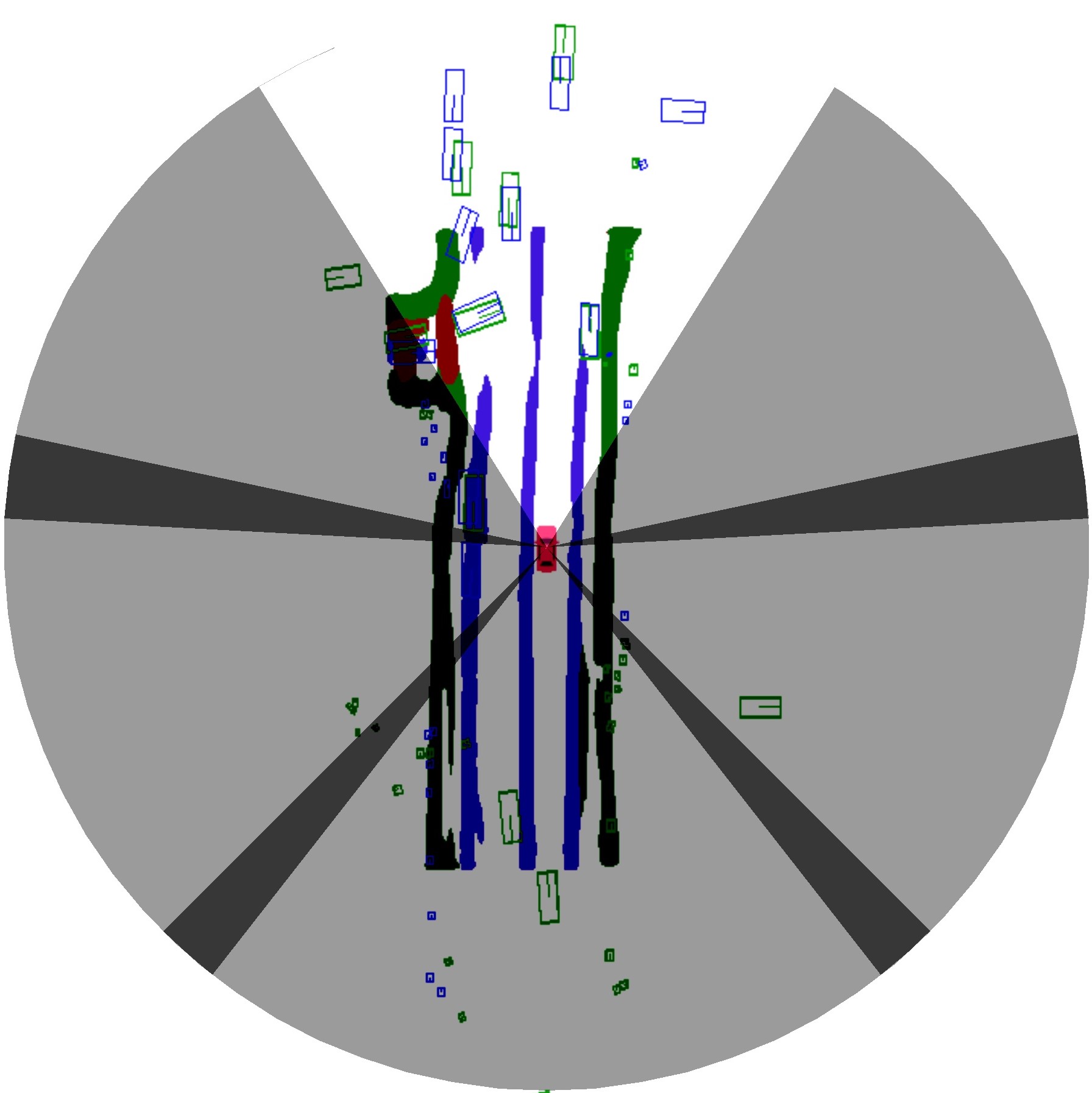}
	\end{subfigure}\\
	\begin{subfigure}{.23\linewidth}
		\centering
		\includegraphics[width=1\linewidth]{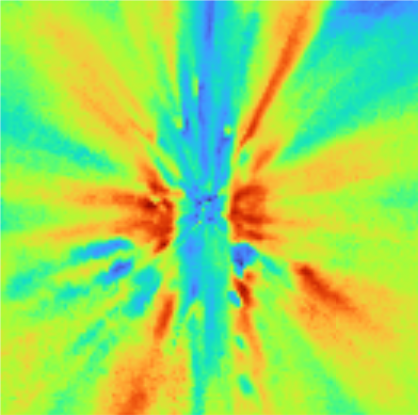}
	\end{subfigure}
	\begin{subfigure}{.23\linewidth}
		\centering
		\includegraphics[width=1\linewidth]{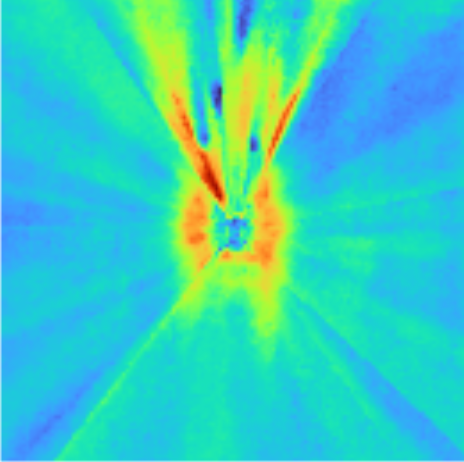}
	\end{subfigure}
	\begin{subfigure}{.23\linewidth}
		\centering
		\includegraphics[width=1\linewidth]{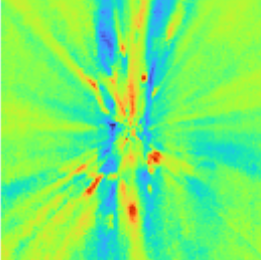}
	\end{subfigure}
	\begin{subfigure}{.23\linewidth}
		\centering
		\includegraphics[width=1\linewidth]{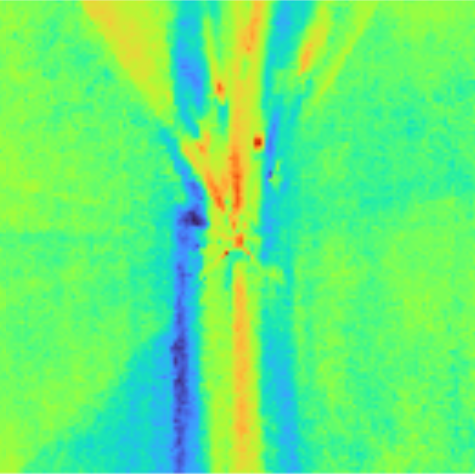}
	\end{subfigure}\\
	\begin{subfigure}{.23\linewidth}
		\centering
		\includegraphics[width=1\linewidth]{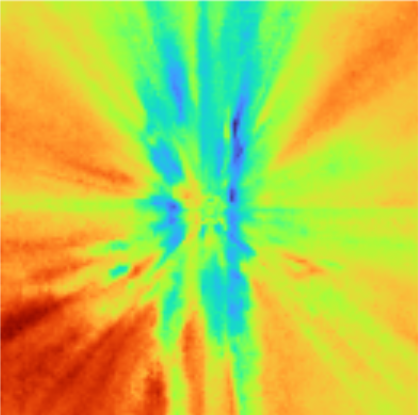}
		\caption{baseline}
		\label{BackBoneFeatures_6imagesvisible2}
	\end{subfigure}
	\begin{subfigure}{.23\linewidth}
		\centering
		\includegraphics[width=1\linewidth]{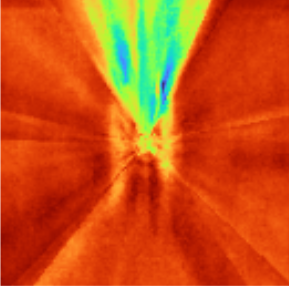}
		\caption{baseline}
		\label{BackBoneFeatures_6images2}
	\end{subfigure}
	\begin{subfigure}{.23\linewidth}
		\centering
		\includegraphics[width=1\linewidth]{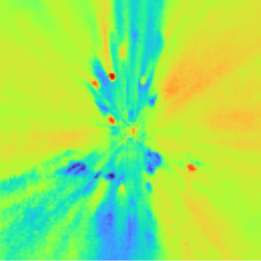}
		\caption{ours}
		\label{BackBoneFeatures_all3_6imagesvisible2}
	\end{subfigure}
	\begin{subfigure}{.23\linewidth}
		\centering
		\includegraphics[width=1\linewidth]{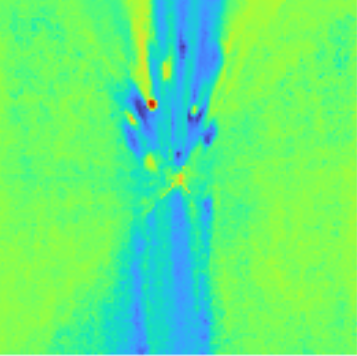}
		\caption{ours}
		\label{BackBoneFeatures_all32}
	\end{subfigure}
	\caption{\ac{BEV} maps and two visualized channels of the latent space BEV feature representation from a six-camera training baseline (a) with six-camera inference and (b) with single-camera inference. In addition, map and features from our training method (c) with six-camera inference (d) with single-camera inference. The gray cover indicates the masked views. The warmer the colors the higher the values.}
	\label{BackBoneFeatures2}
\end{figure}

The effect of our approach in the latent space of the model is visualized in \cref{BackBoneFeatures2}. It shows two of 256 \ac{BEV} feature embeddings. The \ac{BEV} feature embeddings already include the information from the spatial and temporal attention. The features are visualized during inference for both the six camera baseline (\cref{BackBoneFeatures_6images2,BackBoneFeatures_6imagesvisible2}) and our method (\cref{BackBoneFeatures_all32,BackBoneFeatures_all3_6imagesvisible2}). Each training is shown one time with six camera inference and one time with the single camera inference. Even though the feature visualizations are limited in their interpretability, some differences stand out: The \ac{BEV} embeddings of our method (\cref{BackBoneFeatures_all32}) show a hint of the traffic scene even in blind areas as the shape of the street and some objects are more visible compared to the baseline embeddings (\cref{BackBoneFeatures_6images2}). Moreover, our method  (\cref{BackBoneFeatures_all32}) shows more similarity to its six camera equivalent (\cref{BackBoneFeatures_all3_6imagesvisible2}) than the baseline (\cref{BackBoneFeatures_6images2}) to its equivalent (\cref{BackBoneFeatures_6imagesvisible2}). Since both, our and the baseline run have the same single camera input it could only predict more feature information for blind areas by attending into past frames. This richer feature information underlines the more precise results of our run in \cref{VisualResults_sample16}. Additionally, the baseline (\cref{BackBoneFeatures_all3_6imagesvisible2}) shows artificial star-shaped rays which might lead to the reprojection function. In this case, this function might just transport the noised mask fed from the backbone features. These rays are also discussed in \cite{li_fb-bev_2023}. 
\subsection{Effects of features and combinations}
As shown in \cref{ablation_study} the NDS is improved most significantly when the feature reconstruction loss is introduced. The mIoU behaves in the same way. It might be the case that learning what is behind the mask from a completely noise-free sample helps to focus more on the temporal information. The provided sample (\cref{VisualResults_sample16}) underlines the behavior of the values in \cref{ablation}. In more detail, the bounding boxes appear more accurate and show fewer false positives compared to the one-camera baseline.\\
The mAP, which is more influenced by false positive values, drops due to the inverse block masking and the cyclic \ac{LR} which might be the case due to the change in data distribution and reduced input information. This can be improved by the combination of the unmasked sample in the feature reconstruction loss and the ability of larger training steps in the cyclic \ac{LR}. The \ac{GT} for computing the mIoU of the semantic segmentation map is not masked in the blind areas. Since it describes only the prediction of static classes in the \ac{BEV} map, it theoretically has the chance of predicting things like lanes behind the vehicle purely out of past frame information. 
As the results show, this seems to be harder than just guesswork which seems to be the case for the baseline on one camera in \cref{all3vsBase}. It already has better mIoU but shows the most hallucinative visible results as the representative example (\cref{img:16_baseline1image}) underlines. Again the effect of the feature reconstruction loss and thus having a guidance seems to have the most increase in performance to the mIoU. This can be underlined by the latent visualization of \cref{BackBoneFeatures2}. Since the feature reconstruction loss directly impacts the \ac{BEV} feature embedding which changes visibly and needs to rely more on temporal information.
\subsection{Limits}
Due to time and computational constraints, we just developed and tested our training method on the BEVFormer which was trained only on the nuScenes dataset. In addition, our tests were focused on quality improvement, but there is potential for a reduction in computational overhead, as the backbone only needs to be run for one image rather than six at inference time. Even though the method just requires the front camera view during inference it still needs all the \ac{GT} data for the complete sensor setup during training. To determine how the method can be generalized to other models and datasets, as well as to investigate the computational effort and expenses in \ac{GT} data, further investigation is required.
\subsection{Conclusion}
To summarize, our method reduces the number of input images during training for a single camera inference using the BEVFormer model. It reduces the performance degradation, resulting in fewer false-positive detections and more accurate \ac{BEV} segmentation compared to the presented baselines. Additionally, it improves the three most important metrics by $20\%$ NDS, $25\%$ mAP and $19\%$ mIoU.
\bibliographystyle{IEEEtran}
\bibliography{references}

\end{document}